\documentclass[11pt]{article}
\PassOptionsToPackage{table}{xcolor} %
\usepackage[preprint]{acl}

\usepackage{times}
\usepackage{latexsym}
\usepackage[T1]{fontenc}
\usepackage[utf8]{inputenc}
\usepackage{microtype}
\usepackage{inconsolata}
\usepackage{amsmath,amssymb,amsthm}
\usepackage{booktabs}
\usepackage{array}
\usepackage{xspace}
\usepackage{multirow}
\usepackage{graphicx}
\usepackage{url}
\usepackage{pifont}
\usepackage{xfrac}       %
\definecolor{hrblue}{HTML}{3A72B8}\definecolor{hramber}{HTML}{C58B3F}\definecolor{hrrose}{HTML}{C8586B}\definecolor{hrteal}{HTML}{159584}

\definecolor{phyellow}{RGB}{255,242,110}
\definecolor{phedge}{RGB}{176,150,0}
\definecolor{phred}{RGB}{150,30,30}
\definecolor{prorange}{RGB}{252,214,160}
\newcommand{\phid}[1]{\textsuperscript{\tiny\textbf{\textcolor{phred}{[#1]}}}}

\makeatletter
\def\phs@loop#1 #2\@nil{%
  \colorbox{phyellow}{\strut #1\hspace{0.28em}}\allowbreak
  \ifx\relax#2\relax\else\phs@loop#2\@nil\fi}
\DeclareRobustCommand{\phs}[2][?]{{\setlength{\fboxsep}{0pt}\phs@loop #2 \@nil}\phid{#1}}
\makeatother

\newtheorem{proposition}{Proposition}
\theoremstyle{definition}
\newtheorem{definition}{Definition}
\newcounter{finding}

\newcommand{\takeaway}[1]{\par\vspace{3pt}\noindent{\setlength{\fboxsep}{4pt}\setlength{\fboxrule}{0.6pt}%
  \fcolorbox{hrblue}{hrblue!8}{\parbox{\dimexpr\columnwidth-2\fboxsep-2\fboxrule\relax}{%
  \textbf{Takeaway.} #1}}}\par\vspace{3pt}}

\newcommand{\pckci}{PC-KCI}
\newcommand{\heuresis}{\textsc{Heuresis}\xspace}
\newcommand{\dcol}[2]{\begin{tabular}[b]{@{}c@{}}#1\\#2\end{tabular}}   %

\definecolor{chkgreen}{RGB}{200,240,200}
\makeatletter
\def\chk@loop#1 #2\@nil{%
  \colorbox{chkgreen}{\strut #1\hspace{0.28em}}\allowbreak
  \ifx\relax#2\relax\else\chk@loop#2\@nil\fi}
\DeclareRobustCommand{\chk}[1]{{\setlength{\fboxsep}{0pt}\chk@loop #1 \@nil}\textsuperscript{\tiny\textbf{\textcolor{phred}{[CHECK]}}}}
\makeatother

\title{Benchmarking and Enhancing Agentic Causal Discovery}

\author{Pingchuan Ma \\
  Zhejiang University \\ of Technology \\
  \texttt{pma@zjut.edu.cn} \\\And
  Rui Ding \\
  Microsoft \\ \phantom{x} \\
  \texttt{juding@microsoft.com} \\\And
  Shuai Wang \\
  The Hong Kong University \\ of Science and Technology \\
  \texttt{shuaiw@cse.ust.hk}}

\begin{document}
\maketitle

\begin{abstract}
LLM agents can now run causal discovery algorithms on a dataset and report what
comes out. Whether they do more than that, such as noticing where an algorithm
went wrong and fixing it with further tests, is hard to tell from current
benchmarks. Most of them give the variables meaningful names, and an LLM shown
only the names of the Tübingen cause-effect pairs already gets 99\% of the
directions right. With the names removed, the data are usually clean enough that
picking a better algorithm than PC closes most of the gap. We introduce
\heuresis{}, a benchmark of anonymous simulated datasets whose mechanisms range
from linear Gaussian to heavy-tailed. For each family of datasets it separates
what a better existing algorithm can gain from what only a closer look at the data
can gain, and it traces every edge an agent changes. Every family leaves a gap
that none of the algorithms, the router and the two causal-discovery foundation models we ran can close, so a score above the
best chosen algorithm has to come from using the data. We instantiate the benchmark with an agent harness and three LLMs, GPT-6.1 Sol,
Claude Opus 5.5 and Qwen 3.8 Flash. GPT-6.1 Sol surpasses the best chosen
algorithm on all six families, by 14\% adjacency F1 on the most difficult one, and Claude Opus 5.5 does so by 7.4\% there.
A third of the true edges the two agents restore on the two hardest families
appear in no algorithm's output. The code and data will be publicly
released to facilitate future work.
\end{abstract}

\begin{figure*}[t]
\centering
\includegraphics[width=\textwidth]{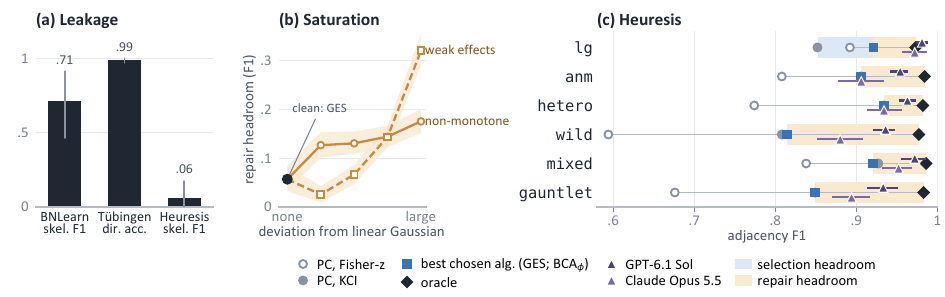}
\caption{{Overview of \heuresis.}
}
\label{fig:chain}
\end{figure*}

\section{Introduction}
\label{sec:intro}

Early work asked LLMs for causal relations from the names of the variables
\citep{kiciman2023causal}, which tests what a model has read rather than whether
it can find a structure nobody knows yet. The setting that matters in practice is
an LLM agent that receives a dataset and carries out causal discovery itself, by
running statistical tests, discovery algorithms and code on the data.
Causal-Copilot \citep{wang2025causalcopilot} picks, tunes and interprets an
algorithm; \citet{zheng2026agents} argue that agents should coordinate the tools
while the causal claims stay grounded in data, and \citet{wu2025llm} that LLMs
should stay out of the decisions altogether.

In this setting the agent decides at several granularities: which
\emph{algorithm} to run, such as PC, GES or LiNGAM; which \emph{component} to use
inside it, such as the conditional-independence (CI) test of PC; which
\emph{local} steps of a run to revisit, such as a test, a conditioning set or a
single deletion; and how to revise the \emph{graph} with further evidence. At the
finest grain the LLM may even judge a dependence itself, in place of a small
tool. The levels form no strict hierarchy, and each can be driven by the data: an
agent that switches to a kernel test because the relations look nonlinear uses
evidence as much as one that repairs an edge. A conventional pipeline fixes these
decisions before it sees the data, whereas an agent can revisit any of them. We
call this capability \emph{fine-grained, evidence-driven adaptation}. Take PC
\citep{spirtes2000causation}, which deletes an edge as soon as one CI test misses
a dependence and never reconsiders the deletion. When a true edge is lost this
way, can the agent find the test, re-test the pair and restore the edge?

Current benchmarks cannot tell, for three reasons. \ding{192}
\textbf{\emph{Leakage}}. Most LLM evaluations of causal discovery name the
variables~\citep{kiciman2023causal, zhou2024causalbench,
babakov2025causalgraphbench}: LLMs reproduce graphs such as Asia and Cancer of
the BNLearn benchmarks from the names alone \citep{srivastava2025realizing}, and
given no data at all an LLM gets 99\% of the Tübingen directions right
(Figure~\ref{fig:chain}a). \ding{193} \textbf{\emph{Saturation}}. With the names
removed, the data are usually clean enough that choosing the algorithm settles
almost everything: on linear-Gaussian data the best of PC and GES is only
$\sim 5\%$ below an ideal oracle that restores every true edge PC wrongly deleted
(Figure~\ref{fig:chain}b), which leaves the finer decisions no room to show.
\ding{194} \textbf{\emph{Opacity}}. Existing benchmarks score only the final
graph, so they cannot tell whether a gain came from a better existing algorithm
or from analysis inside the run. Agents may well make fine-grained decisions
already; what is missing is a measurement that credits a gain to them.

\heuresis{} measures this adaptation without leakage, saturation or opacity. Its
datasets are anonymous and drawn from six families of mechanisms, from linear
Gaussian to heavy-tailed, so that substantial headroom remains after the
algorithm is chosen. For every family it fixes reference points before any agent
is scored: the best algorithm chosen globally and per family, and an ideal
oracle that repairs every wrong deletion of PC (Figure~\ref{fig:chain}c).
Choosing per family gains nothing on five of the six families, where PC with a
kernel CI test (KCI) is already best, while the gap to the oracle stays
substantial on every family and is largest where every algorithm scores lowest.
We probe the finer levels through the one decision
type whose correctness the ground truth fixes exactly, PC's edge deletions: each is logged with its
test and conditioning set, the ground truth says which were wrong, and every edge
an agent restores can be traced to the calls before it. The benchmark therefore
reports the gain over existing algorithms and selection strategies separately
from the behavior in the trajectory that produced it.

We run a tool-enabled agent harness with GPT-6.1 Sol, Claude Opus 5.5 and
Qwen~3.8~Flash, under one toolbox and budget and without telling them the family.
GPT-6.1 Sol improves the best chosen algorithm on every family, by 14\% on
\texttt{wild}, the most difficult family, and Claude Opus 5.5 by 7.4\% there;
Qwen~3.8~Flash reaches the best chosen algorithm on linear-Gaussian data.
Their traces show that no $k$-vote merge of the algorithms' outputs, with $k$ chosen
per family in hindsight, reproduces these gains, and that a third of the true edges they restore on the two hardest
families appear in no algorithm's output at all. Our
contributions are:
\begin{itemize}\itemsep0em
\item We frame agentic causal discovery as evidence-driven decisions at several
granularities and show that leakage, saturation and opacity keep existing
benchmarks from crediting a gain to the decisions behind it.
\item We propose \heuresis, anonymous simulated data with headroom beyond
algorithm choice, reference points for choosing and for repairing, a traced
probe on PC's deletions, and an agent harness that records every call.
\item We find that two of the three LLMs surpass the best chosen
algorithm by re-testing what PC deleted, and trace the edges they restore to the
calls that produced them.
\end{itemize}

\section{Related Work}
\label{sec:related}

\paragraph{LLMs and causal discovery.}
\citet{kiciman2023causal} show that LLMs recover causal directions from variable
names alone. Corr2Cause and CLadder remove the names but pose closed-form
questions with no data \citep{jin2024corr2cause, jin2023cladder}, while
CausalBN-Bench and CausalGraphBench compare LLMs with classical algorithms on
named BNLearn graphs \citep{zhou2024causalbench, babakov2025causalgraphbench},
which LLMs have largely memorized \citep{srivastava2025realizing,
yu2026caliper}. A second line uses the LLM as a component of a classical
algorithm, for instance to answer or prioritize its independence tests from the
variable names \citep{cohrs2024chatpc, hiremath2025guess2graph}, or to check and
orient the edges it learned \citep{ban2023ilscsl, khatibi2024alcm,
lyu2025mosacd}. These works primarily evaluate the causal knowledge of LLMs
w.r.t.~the variable names instead of purely statistics-driven causal discovery.

\paragraph{Agents for causal and statistical analysis.}
Causal-Copilot and CausalAgent let an LLM run a whole analysis
\citep{wang2025causalcopilot, han2024causalagent}, and QRData, DSBench,
CauSciBench and P-Bench score the final answer on real data
\citep{liu2024qrdata, jing2025dsbench, acharya2026causcibench,
miao2026fisherr1}. A final answer cannot show whether the agent chose a better
method or used the evidence, and agent scores move with the budget and the
harness \citep{kapoor2024agents, kapoor2025hal}.

\paragraph{Simulated discovery benchmarks for agents.}
CausalDS, CausaLab and CausalArena generate data from hidden causal graphs and
score the graph an agent recovers \citep{leban2026causalds, yang2026causalab,
li2026causalarena}; CausalArena also finds that which algorithm works best
depends on the kind of data. The closest benchmark is ACDB \citep{deb2026acdb},
which gives LLM agents anonymized simulated data and compares them with PC. Its
data are simply linear Gaussian with few variables and it scores only the final
graph.

\section{Background}
\label{sec:background}

\paragraph{Causal discovery from data.}
A causal discovery algorithm takes a table of observations over $d$ variables
and returns a graph whose edges mean direct influence; the true graph is a
directed acyclic graph (DAG). We score an output graph mainly on its
\emph{skeleton}, the edges without directions, by \emph{adjacency F1}, the F1
score of the output edges against the true edges. Directions are scored by the
\emph{structural Hamming distance} (SHD), the number of edge edits to the true
graph.

\begin{figure}[t]
\centering
\includegraphics[width=\columnwidth]{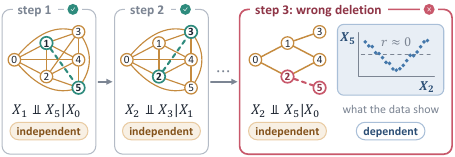}
\caption{Running example of PC.}
\label{fig:pc}
\end{figure}

\paragraph{PC and wrong deletions.}
PC \citep{spirtes2000causation} is the standard algorithm in this paper
(Figure~\ref{fig:pc}). It
starts from a fully connected graph and removes the edge between two variables
when a statistical test, a Fisher-$z$ test or a kernel-based conditional
independence (KCI) test \citep{zhang2011kernel}, declares them independent given
some other variables, the \emph{conditioning set}. PC reaches the data only
through this test, and a deletion is final: when the test misses a real
dependence, a true edge is lost. These \emph{wrong deletions} are the primary
source of PC's errors. PC records every deletion with the test and the
conditioning set behind it, the \emph{deletion log}. We focus on whether agents
can repair wrong deletions, the most direct way to use the evidence to improve
the output.

\paragraph{Example.} Figure~\ref{fig:pc} gives a running example. PC starts from the complete graph and deletes an edge (dashed) when
a test finds its two variables independent given a set $S$ of others. In steps 1
and 2 the hidden graph agrees (\checkmark). In step 3 it does not: $X_5$ depends
on $X_2$ through a V-shaped map, which a Fisher-$z$ test, blind to non-linear
dependence, reads as independence. The true edge is lost and never revisited.

\paragraph{Other algorithms and the ideal oracle.}
GES \citep{chickering2002optimal} searches over graphs with a score instead of
tests and suits linear-Gaussian data; FCI, LiNGAM \citep{shimizu2006lingam},
additive-noise methods \citep{hoyer2008nonlinear} and NOTEARS rest on other
assumptions. Real data violate many of these assumptions, so no single algorithm
is best. The \emph{ideal oracle}, which restores every wrongly deleted edge of
PC, is the ceiling for repairing deletions, not for the task: an agent can also gain
outside PC's deletions (\S\ref{sec:reachable}).

\begin{figure*}[t]
\centering
\includegraphics[width=\textwidth]{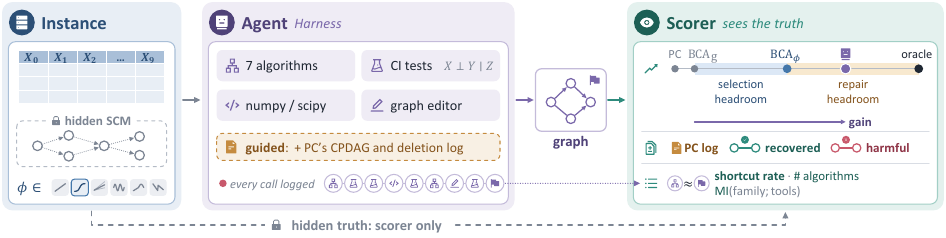}
\caption{A \heuresis{} episode and its evaluation.}
\label{fig:task}
\end{figure*}

\begin{table}[t]
\centering
\footnotesize
\setlength{\tabcolsep}{4pt}
\renewcommand{\arraystretch}{1.15}
\resizebox{\columnwidth}{!}{%
\begin{tabular}{lcccc}
\toprule
Benchmark & \dcol{D1}{Anonymous} & \dcol{D2}{Headroom} & \dcol{D3}{Gains} & \dcol{D4}{Decomposed} \\
\midrule
Corr2Cause         & \checkmark & $\times$   & $\times$   & $\times$ \\
CausalBN-Bench     & $\times$   & $\times$   & $\times$   & $\times$ \\
chatPC             & $\times$   & $\times$   & $\times$   & $\sim$ \\
CausalArena        & --         & $\times$   & $\times$   & $\times$ \\
ACDB               & \checkmark & $\times$   & $\times$   & $\times$ \\
\midrule
\heuresis{} (ours) & \checkmark & \checkmark & \checkmark & \checkmark \\
\bottomrule
\end{tabular}}
\caption{Existing benchmarks against the design requirements D1--D4, as described in their papers. \checkmark{} yes, $\sim$ partly, $\times$ no,
-- not reported.}
\label{tab:compare}
\end{table}

\begin{table*}[t]
\centering
\footnotesize
\setlength{\tabcolsep}{2.6pt}
\renewcommand{\arraystretch}{1.2}
\newcommand{\famon}{\textcolor{hrblue}{\ding{108}}}%
\newcommand{\famoff}{\textcolor{black!20}{\textperiodcentered}}%
\newcommand{\fambar}[1]{#1~\textcolor{hramber!70}{\rule{#1\dimexpr110pt\relax}{1.3ex}}}%
\begin{tabular}{@{}l c ccc cc ccc c @{\hspace{10pt}} r r l@{}}
\toprule
 & & \multicolumn{3}{c}{\emph{nonlinear}} & \multicolumn{2}{c}{\emph{input-dep.\ noise}}
 & \multicolumn{3}{c}{\emph{non-monotone}} & \emph{$t_3$ noise} & \multicolumn{3}{c}{difficulty} \\
\cmidrule(lr){3-5}\cmidrule(lr){6-7}\cmidrule(lr){8-10}\cmidrule(lr){11-11}\cmidrule(l){12-14}
Family & linear & MLP & piecewise & sigmoid & MLP & sigmoid & quadratic & $|\cdot|$ & Fourier & linear & Wrong & Zero & Headroom \\
\midrule
\texttt{lg}       & \famon  & \famoff & \famoff & \famoff & \famoff & \famoff & \famoff & \famoff & \famoff & \famoff & 2.4 & 23\% & \fambar{.121} \\
\texttt{anm}      & \famoff & \famon  & \famoff & \famoff & \famoff & \famoff & \famoff & \famoff & \famoff & \famoff & 1.4 & 26\% & \fambar{.078} \\
\texttt{hetero}   & \famoff & \famoff & \famoff & \famoff & \famon  & \famoff & \famoff & \famoff & \famoff & \famoff & 1.0 & 39\% & \fambar{.049} \\
\addlinespace[2pt]
\texttt{wild}     & \famon  & \famon  & \famoff & \famoff & \famon  & \famoff & \famon  & \famon  & \famon  & \famon  & 3.0 & 5\% & \fambar{.169} \\
\texttt{mixed}    & \famon  & \famon  & \famoff & \famoff & \famon  & \famoff & \famoff & \famoff & \famoff & \famoff & 1.2 & 36\% & \fambar{.059} \\
\texttt{gauntlet} & \famoff & \famoff & \famon  & \famon  & \famoff & \famon  & \famon  & \famon  & \famon  & \famoff & 2.4 & 9\% & \fambar{.134} \\
\bottomrule
\end{tabular}
\caption{Mechanism families of \heuresis}

\label{tab:families}
\end{table*}

\section{\texorpdfstring{\heuresis{}}{Heuresis}}
\label{sec:bench}

The three failures of \S\ref{sec:intro} lead to four design requirements of
\heuresis. \emph{\textbf{D1:} anonymous variables.} Nothing can be recalled from
a name. \emph{\textbf{D2:} both kinds of headroom measured per family.} For
every family, the benchmark measures how much of the gap above the global best chosen
algorithm a better existing algorithm closes and how much no existing algorithm
or the router recovers, before any agent is scored. \emph{\textbf{D3:}
one gain per reference.} Gains over the global best chosen algorithm and over the best chosen algorithm of the
family must separate an agent that only calls the right algorithm, one that edits the graph blindly and one
that improves on the right algorithm. \emph{\textbf{D4:} every graph decomposed and every call
recorded}, so that a score can be explained without further experiments.
Table~\ref{tab:compare} sets existing benchmarks against these four
requirements.

\subsection{Task and mechanism families}
\label{sec:task}
\label{sec:gen}
\begin{figure}[t]
\centering
\includegraphics[width=\columnwidth]{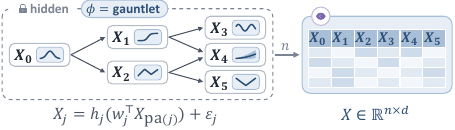}
\caption{Running example of \heuresis instance.}
\label{fig:space}
\end{figure}

We separate three levels: a mechanism family $\phi$, the SCMs $S$ drawn from it,
and the data $X$ sampled from one SCM.
\begin{definition}[SCM]
  A \emph{mechanism} $m=(f,\nu)$ pairs a function $f$ with a noise distribution
  $\nu$. A \emph{structural causal model} (SCM) $S=\bigl(G,(m_j)_{j=0}^{d-1}\bigr)$
  consists of a DAG $G$ over $d$ variables and a mechanism $m_j=(f_j,\nu_j)$ for
  each variable, which generates the variables as
  \begin{equation}
    X_j = f_j\bigl(X_{\mathrm{pa}_G(j)}, \varepsilon_j\bigr), \quad
    \varepsilon_j \sim \nu_j,
  \end{equation}
  for $j = 0, \ldots, d-1$, where the noise terms $\varepsilon_j$ are
  independent and $\mathrm{pa}_G(j)$ denotes the parents of $j$ in $G$. Drawing $n$ independent samples of
  $(X_0,\ldots,X_{d-1})$ from $S$ gives a data matrix
  $X \in \mathbb{R}^{n \times d}$.
\end{definition}

An \emph{instance} of \heuresis{} is the data matrix $X$ of a hidden SCM $S \sim
\mathcal{S}_\phi$, with $S$ and its DAG $G$ kept as the ground truth
(Figure~\ref{fig:space}). The agent sees only $X$, whose columns are anonymous
and in random order. In an episode, the agent is asked to recover $G$ from $X$.
As one can expect, the difficulty of the instance largely depends on the family
$\phi$ from which $S$ is drawn, and we define it as follows.
\begin{definition}[Mechanism family]
  A \emph{mechanism family} $\phi$ is a distribution over mechanisms. An SCM of
  family $\phi$, written $S \sim \mathcal{S}_\phi$, is drawn in two steps:
  first, the DAG $G$ is drawn from a random-graph prior
  (Appendix~\ref{app:datasheet}), and then each variable draws its own mechanism
  $m_j \sim \phi$ independently of the others, with $f_j$ taking as input the
  $|\mathrm{pa}_G(j)|$ parents of $j$.
\end{definition}
In \heuresis, we introduce six families of mechanisms, from linear Gaussian to
heavy-tailed, as summarized in Table~\ref{tab:families}. The families are
designed to stress the assumptions of the standard algorithms, and they are
ordered by which assumption each family removes, not by difficulty. The first
three families (\texttt{lg}, \texttt{anm} and \texttt{hetero}) are
single-mechanism families, where every variable draws its mechanism from the
same distribution (i.e., linear Gaussian, MLP mean, or MLP mean and noise scale,
respectively) . The last three families (\texttt{wild}, \texttt{mixed} and
\texttt{gauntlet}) are mixtures, where every variable draws its mechanism from a
different distribution. In particular, we deem \texttt{wild} the most difficult
family, because it involves highly sophisticated mechanisms that are nonlinear,
non-monotone and heavy-tailed. In \texttt{wild}, every variable can have a
different mechanism and the assumptions of the standard algorithms fail in
multiple ways. \texttt{gauntlet} draws each variable from the widest set of nonlinear,
non-monotone and input-dependent-noise mechanisms (Table~\ref{tab:families}), so
that a single instance breaks several assumptions at once. The benchmark
is designed to be extensible, so that new families can be added in the future.
We report the full details of these mechanism families in
Appendix~\ref{app:datasheet}.

\subsection{Agent harness}
\label{sec:harness}
\begin{figure}[t]
\centering
\includegraphics[width=\columnwidth]{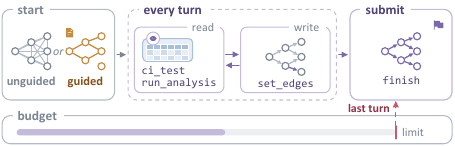}
\caption{The working graph over an episode.}
\label{fig:harness}
\end{figure}

An episode is an iterative edit of a \emph{working graph}
(Figure~\ref{fig:harness}). The agent starts from an initial graph and, in each
turn, reads evidence from the data and writes changes to the graph, until it
submits. The harness is built around two requirements: evidence is available on
demand, and each decision can be traced back to the evidence behind it.

\paragraph{Interaction.}
The state of an episode is the working graph, which assigns each pair of variables
one relation: no edge, an undirected edge, or an edge in either direction. In each
turn the agent takes one of four kinds of actions: it runs a structure-learning
algorithm, queries evidence, analyzes the data, or edits and submits the graph. The
observations are the structured results of these tools, and the submitted graph is
completed and scored as in \S\ref{sec:labels}. The unit of decision is a single
variable pair rather than the whole graph, which is what makes the repair headroom
$H^{\mathrm{r}}_\phi$ reachable by an agent at all.

\paragraph{Evidence at three granularities.}
The toolbox offers evidence at three levels. \ding{192} At the \emph{graph
level}, a menu of structure-learning algorithms returns a complete graph per
call: PC with a Fisher-$z$ test and with KCI, FCI, GES, DirectLiNGAM, an
additive-noise method and linear NOTEARS \citep{zheng2024causallearn}. Choosing
among them is the selection headroom $H^{\mathrm{s}}_\phi$ of
\S\ref{sec:labels}, which most existing causal discovery agents target.
\ding{193} At the \emph{edge level}, the agent can run a
conditional-independence test, Fisher-$z$ or KCI \citep{zhang2011kernel,
strobl2019approximate}, with a conditioning set of its own choice, and check how
stable a verdict is under resampling of the data. This is the kind of evidence
PC consumes when it deletes an edge, so the agent can repeat, replace or
overturn a deletion on the same footing as the algorithm. \ding{194} At the
\emph{mechanism level}, the agent can summarize the data, inspect how well a
variable is explained by a candidate set of parents, probe the direction of a
pair, and write short numerical programs (\texttt{numpy} and \texttt{scipy})
that run in an isolated process. These tools let it test whether the assumptions
of an algorithm hold on this instance, for instance linearity, a constant noise
scale or light tails, and decide whose output to trust. The menu is thus the
coarse end of the toolbox, and the two finer levels are where the decisions
inside a run are made.

\paragraph{Resources and termination.}
Evidence is charged by what it costs to produce. Calls that re-run a statistical
test draw on a test budget, counted in the same unit as the cost of PC itself, while
diagnostics and analysis are bounded only by computation. The number of turns, the
output per turn, the number of algorithm calls and the total running time are also
limited, and all limits are visible to the agent (values in
Appendix~\ref{app:harness}). When any of them is exhausted, the agent gets one last
turn in which it can only submit. Every episode therefore returns a graph, which is
the precondition for reporting every cell over all episodes (\S\ref{sec:labels},
Appendix~\ref{app:truncation}).

\paragraph{What the agent sees.}
A test returns a verdict with a coarse strength grade, not a raw $p$-value, so that
decisions rest on interpreting the evidence instead of tuning against a threshold.
The identity of each algorithm is masked within an episode, so that the agent
chooses a tool for what the data suggest and not for its name. The default prompt is
neutral: it states neither the metric nor the conditions under which an algorithm
applies (variants in \S\ref{sec:practice}). Analysis code has access to the data
only, not to the ground truth or the state of the harness.

\paragraph{Record.}
The harness records every call with its type, arguments, result and cost. Each edge
of a submitted graph can thus be traced to the edits and the evidence behind it, and
the records are the source of the behavioral statistics in \S\ref{sec:labels} and
the trace-level attribution in \S\ref{sec:results}.

\paragraph{Guided vs. unguided.}
\label{sec:seeded}
The two conditions differ only in the starting point of the working graph and in what
the agent is shown. In the \emph{guided} condition, the working graph starts from the
output graph of \pckci{} and the agent receives the deletion log of \pckci{}: for each
deleted pair, the conditioning set, the test and the strength of the evidence that
removed it, but not the truth label. In the \emph{unguided} condition, the agent
starts from the complete undirected graph and edits it based on its own decisions.
The difference between the two is the joint effect of a good starting point and of
knowing how PC decided, so an agent whose guided and unguided scores coincide
gains from neither. Guided episodes also give the precision and recall of the
agent's repairs per deletion (\S\ref{sec:results}).

\subsection{Evaluation protocol}
\label{sec:labels}

\paragraph{Scores.}
Every submitted graph is first completed with the standard orientation rules
\citep{meek1995causal} and then scored by adjacency F1, the primary score, and
by normalized SHD (nSHD) against the true graph
(\S\ref{sec:background}) as a secondary metric. Adjacency F1 is the primary score
because the skeleton is what observational data identify under faithfulness.
Every episode must submit a graph.
Scoring only the episodes that finish would keep the easy instances and could mimic
a gain beyond algorithm choice (Appendix~\ref{app:truncation}), so we report every
cell over all episodes and claim a capability only when finished episodes agree.

\paragraph{Headroom Estimation.}
Before any agent is scored, we estimate for every family $\phi$ how much a
better algorithm and how much further evidence could add. To this end, we run a
collection of existing algorithms on every instance and select, on development instances, the
global best chosen algorithm (BCA$_{\mathrm g}$), the one with the best mean adjacency F1 across all
families, and the best chosen algorithm for each family (BCA$_\phi$), the one with the best mean on that
family. As introduced in \S\ref{sec:background}, PC deletes edges sequentially and a deletion is final, so we
further define an \emph{ideal oracle} that restores every wrongly deleted edge of PC and leaves the rest of
the output unchanged. With these references, we estimate two kinds of headroom. The \emph{selection headroom}
\begin{equation}
  H^{\mathrm{s}}_\phi = \mathrm{F1}(\mathrm{BCA}_\phi) - \mathrm{F1}(\mathrm{BCA}_{\mathrm{g}}),
\end{equation}
is what choosing the right algorithm for the family gains and is the focus of
many existing causal discovery agents, and the \emph{repair headroom}
\begin{equation}
  H^{\mathrm{r}}_\phi = \mathrm{F1}(\text{oracle}) - \mathrm{F1}(\mathrm{BCA}_\phi),
\end{equation}
is what remains for decisions inside a run, measured on PC's deletions, and is of interest in this paper.
We report both headrooms in our experiments.

\paragraph{Main Metrics.} With the two headrooms defined, we can now define the
metrics that place an agent against these references. Let $\mathrm{F1}(A)$ be
the adjacency F1 of an agent $A$ on a family $\phi$. We define \emph{gain} as
$\Delta^{\mathrm{s}}_\phi(A) = \mathrm{F1}(A) -
\mathrm{F1}(\mathrm{BCA}_{\mathrm{g}})$, which is the improvement of the agent
over the global best chosen algorithm, and \emph{residual gain} as
$\Delta^{\mathrm{r}}_\phi(A) = \mathrm{F1}(A) - \mathrm{F1}(\mathrm{BCA}_\phi)$,
which is the improvement of the agent over the best chosen algorithm for that
family.

\paragraph{Drill-down Analysis.}
Since \heuresis holds the ground truth, it can decompose every submitted graph
into the edges recovered and harmed in specific episodes. A submitted graph is
then compared with the deletion log of PC: the wrongly deleted pairs it contains as
edges (\emph{recovered}), the rightly deleted pairs it contains as edges
(\emph{harmful}), the edges of PC it dropped, and the pairs it re-oriented. Three behavioral statistics come from the tool-call
record: the \emph{shortcut rate}, the share of episodes whose final graph is within
one SHD of the output of the first call to a PC variant; the number of distinct
algorithms called; and \emph{tool-choice dependence}, the mutual information
between the family and the tools called.

\section{Experiments}
\label{sec:validity}

\paragraph{Setup.}
All instances have $d=10$ and $n=600$, with 100 development and 32 test
instances per family. Reference algorithms are chosen on development instances
and every result is on test instances. Then, we instantiate our harness with
three LLMs, GPT-6.1 Sol, Claude Opus 5.5 and Qwen~3.8~Flash,
and run them on the test instances of all six families. The agents are given the
same toolbox and budget, without telling them the family. We report the
adjacency F1 of every agent on every family, against the best chosen algorithm for
that family and the ideal oracle. Intervals in the figures are 95\% paired bootstrap intervals
over instances; the full numbers are in Appendix~\ref{app:full}.

\subsection{Headroom above BCA on \heuresis}
\label{sec:headroom}
\label{sec:recoverable}

\paragraph{Baselines.}
Besides the best chosen algorithm for each family (BCA$_\phi$), we compare three
baselines that claim to adapt to diverse data: \ding{192} \texttt{router}, a
non-LLM adaptive controller that sees the same data and CI tests as the agents, screens the data, routes to a suitable algorithm and
re-tests the deleted pairs (Appendix~\ref{app:baselines}), and the two recent
causal-discovery foundation models \ding{193} \texttt{CDFM}~\citep{qiao2026cdfm}
and \ding{194} \texttt{TabCausal}~\citep{li2026tabcausal}, run as released with
default thresholds and no family label (Appendix~\ref{app:baselines}). We run them on the test
instances of all six families and report their adjacency F1 relative to
BCA$_\phi$ (Table~\ref{tab:headroom}); Appendix~\ref{app:signals} adds scripted
repairs of PC's output, up to a gradient-boosted classifier.

\begin{table}[t]
\centering
\footnotesize
\setlength{\tabcolsep}{3pt}
\resizebox{\columnwidth}{!}{%
\begin{tabular}{lrrrrrr}
\toprule
 & \texttt{lg} & \texttt{anm} & \texttt{hetero} & \texttt{wild} & \texttt{mixed} & \texttt{gauntlet} \\
\midrule
BCA$_\phi$  & .921 & .906 & .934 & .814 & .920 & .849 \\
Oracle                & .973 & .984 & .982 & .977 & .986 & .983 \\
\midrule
$H^{\mathrm{s}}_\phi$    & \cellcolor{hrblue!59}\textbf{.069} & .000 & .000 & .006 & $-$.007 & .000 \\
$H^{\mathrm{r}}_\phi$         & \cellcolor{hramber!18}.052 & \cellcolor{hramber!28}.078 & \cellcolor{hramber!17}.049 & \cellcolor{hramber!58}\textbf{.163} & \cellcolor{hramber!23}.066 & \cellcolor{hramber!47}.134 \\
\midrule
$\Delta^{\mathrm{r}}_\phi(\texttt{router})$ & \cellcolor{hrrose!9}$-$.029 & .000 & .000 & $-$.006 & .007 & .000 \\
$\Delta^{\mathrm{r}}_\phi(\texttt{CDFM})$ & \cellcolor{hrrose!56}$-$.186 & \cellcolor{hrrose!57}$-$.190 & \cellcolor{hrrose!53}$-$.177 & \cellcolor{hrrose!31}$-$.104 & \cellcolor{hrrose!57}$-$.191 & \cellcolor{hrrose!37}$-$.124 \\
$\Delta^{\mathrm{r}}_\phi(\texttt{TabCausal})$ & \cellcolor{hrrose!8}$-$.016 & .001 & \cellcolor{hrrose!14}$-$.046 & \cellcolor{hrteal!8}.019 & \cellcolor{hrrose!8}$-$.028 & \cellcolor{hrteal!9}.029 \\
\bottomrule
\end{tabular}} \caption{F1 of the best chosen algorithm (BCA$_\phi$) and the
oracle, the two headrooms, and the residual gain $\Delta^{\mathrm{r}}_\phi$ of
three baselines. 
}
\label{tab:headroom}
\end{table}

An oracle that restores only the one to three true edges PC wrongly deletes on an
instance, out of roughly ten (Table~\ref{tab:families}), scores .97 to .99 on every
family, whereas the best chosen algorithm lags 5\% to 16\% behind, most on
\texttt{wild} (16\%) and \texttt{gauntlet} (13\%). Non-monotone relations, not
heavy-tailed noise, open the gap (Figure~\ref{fig:chain}b).

Choosing the best algorithm per family gains .07 on \texttt{lg} and nothing
beyond $\pm.007$ elsewhere (Figure~\ref{fig:chain}c). The hand-written router,
\texttt{CDFM} and \texttt{TabCausal} never beat the best chosen algorithm by more
than .03, and \texttt{CDFM} is 10\% to 19\% below it (Table~\ref{tab:headroom}).
Repairing the output of PC from its own log does not help. Where the
deleting test has little power, its recorded $p$-value says almost nothing about
whether the deletion was wrong (Proposition~\ref{prop:limit},
Appendix~\ref{app:proof}), and restoring edges by signals that need new tests
does not raise F1 either (Appendix~\ref{app:signals}). Closing the gap therefore takes
evidence from the individual dataset, which we test with agents in
\S\ref{sec:reachable}.

\takeaway{\heuresis{} leaves room that evidence can fill: an oracle that repairs
PC's wrong deletions scores .97 to .99 on every family, the best chosen algorithm
lags 5\% to 16\% behind, and none of the algorithms, the router, the two foundation
models and the scripted repairs we ran closes the gap.}

\subsection{Agent Gains over the Best Chosen Algorithm}
\label{sec:reachable}

\begin{figure}[t]
\centering
\includegraphics[width=\columnwidth]{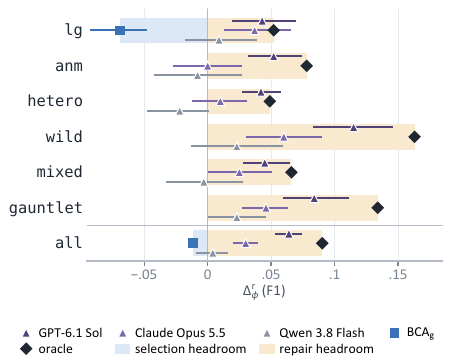}
\caption{Residual gain on \heuresis of LLM agents.}
\label{fig:agents}
\end{figure}

Figure~\ref{fig:agents} shows $\Delta^{\mathrm{r}}_\phi$ of each agent on every
family, against the two kinds of headroom (full results in Table~\ref{tab:m1},
Appendix~\ref{app:full}). GPT-6.1 Sol improves the best chosen algorithm on all
six families, and its largest gain, 14\%, is on \texttt{wild}, the most
difficult family. On \texttt{lg} its .978 lies above the oracle's .973, since it also gains
outside PC's deletions. Claude Opus 5.5 improves the best chosen algorithm on
\texttt{wild} by 7.4\%, about a third of that family's repair headroom, is on
par with the ideal oracle on \texttt{lg}, and loses on no family. Qwen~3.8~Flash
reaches the best chosen algorithm on \texttt{lg} without being told the family,
which closes the selection headroom, but it stops there and closes none of the
repair headroom. It is on par with \pckci{} on \texttt{anm}, \texttt{wild} and
\texttt{mixed}, slightly below it on \texttt{hetero}, and recovers only a small
part of the headroom above \pckci{} on \texttt{gauntlet}. Over all families (row
all), the $\Delta^{\mathrm{r}}_\phi$ of GPT-6.1 Sol is about twice that of Claude
Opus 5.5. Repeating
GPT-6.1 Sol and Claude Opus 5.5 three times on \texttt{lg}, \texttt{anm},
\texttt{hetero} and \texttt{wild} leaves the ordering of the two agents, and the
families on which $\Delta^{\mathrm{r}}_\phi$ lies above zero, unchanged
(Table~\ref{tab:m1rep}).
GPT-6.1 Sol earns this gain at about \$0.19 per episode, twice the gain of Claude
Opus 5.5 at an eighth of its cost (Table~\ref{tab:harness}). On the DAG target both
agents also lower the nSHD of PC-KCI's output on every family
(Table~\ref{tab:nshd}).
\takeaway{LLM-based agents achieve positive residual gains
$\Delta^{\mathrm{r}}_\phi$ on \heuresis{}: the mean $\Delta^{\mathrm{r}}_\phi$
of every agent is above zero over all families, and on all six families for
GPT-6.1 Sol, five for Claude Opus 5.5 and three for Qwen~3.8~Flash.}

\begin{table}[t]
\centering
\footnotesize
\setlength{\tabcolsep}{2.5pt}
\resizebox{\columnwidth}{!}{%
\begin{tabular}{lrrrrrrrr}
\toprule
 & \multicolumn{4}{c}{GPT-6.1 Sol} & \multicolumn{4}{c}{Claude Opus 5.5} \\
\cmidrule(lr){2-5}\cmidrule(lr){6-9}
 & \pckci{} & merge$^*$ & agent & unsupp. & \pckci{} & merge$^*$ & agent & unsupp. \\
\midrule
\texttt{lg}       & .854 & .892 & .978 & 0/72  & .838 & .884 & .972 & 8/61 \\
\texttt{anm}      & .887 & .887 & .958 & 4/39  & .891 & .891 & .910 & 2/29 \\
\texttt{hetero}   & .914 & .920 & .967 & 5/29  & .913 & .922 & .940 & 3/24 \\
\texttt{wild}     & .813 & .813 & .935 & 20/57 & .817 & .817 & .885 & 13/40 \\
\texttt{mixed}    & .919 & .933 & .972 & 4/34  & .915 & .928 & .951 & 0/28 \\
\texttt{gauntlet} & .835 & .835 & .933 & 15/47 & .833 & .833 & .895 & 7/32 \\
\bottomrule
\end{tabular}} \caption{Trace attribution. \emph{merge$^*$} is the best F1 reached by adding to
the \pckci{} output every edge that at least $k$ of the other algorithms the agent
called propose, with $k$ chosen per family in hindsight. \emph{Unsupp.}: true
edges missing from the \pckci{} output that the agent restores although no
algorithm it called proposes them, out of all such edges it restores.}
\label{tab:trace}
\end{table}

\subsection{Trace-Level Attribution}
\label{sec:results}

To see how the agents reach their gains, we trace every edge they add to the
\pckci{} output they obtained (Table~\ref{tab:trace}).

A straightforward hypothesis is that the agents merge the outputs of several
algorithms, and that the edges they restore are those that at least $k$ of the
algorithms propose, with $k$ chosen per family in hindsight. The traces rule this
out. As shown in Table~\ref{tab:trace}, even the best $k$-vote merge of the
outputs the agent called falls far below the agent's own score on \heuresis.
Over \emph{merge$^*$}, for instance, GPT-6.1 Sol gains 8.9\% on average. On
\texttt{wild} and \texttt{gauntlet}, where the merge gains nothing over \pckci{},
35 of the 104 true edges that GPT-6.1 Sol restores, about a third, appear in no
algorithm's output.

\begin{figure}[t]
\centering
\includegraphics[width=\columnwidth]{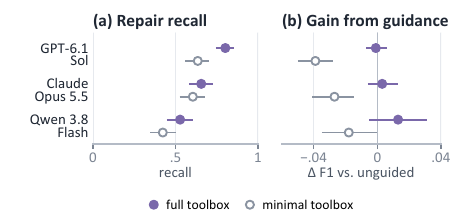}
\caption{(a) Recall of the wrongly deleted edges. (b) Gain in F1 of the guided
episode over the unguided episode on the same instance. }
\label{fig:guided}
\end{figure}

As a second hypothesis, the agents could be reading the deletion log of \pckci{}
and restoring the edges it wrongly deleted. We run the guided condition with the full toolbox and with a minimal toolbox
that lacks the code tool, the stability test and the orientation probes. The results are reported in
Figure~\ref{fig:guided} (see full results in Table~\ref{tab:guided},
Appendix~\ref{app:full}). We observe that GPT-6.1 Sol, the best-performing LLM,
restores nearly all of the wrongly deleted edges on \texttt{lg}. Handing them both the output of PC and its
deletion log adds nothing for GPT-6.1 Sol and Claude Opus 5.5 (the interval of
the guided minus unguided gain contains zero on every family), so neither channel
accounts for their gains. The toolbox does matter: without the code tool, the
stability test and the orientation probes, GPT-6.1 Sol's recall on \texttt{wild}
falls from .70 to .47.

\takeaway{GPT-6.1 Sol and Claude Opus 5.5 go beyond BCA$_\phi$ through evidence
they gather themselves: they test individual edges and probe the mechanism with
code, and neither a merge of the algorithms' outputs nor the output and deletion
log of PC reproduces the gain.}

\subsection{Robustness to Harness Settings}
\label{sec:practice}

The residual gains do not depend on how the harness is set. On \texttt{lg}, the
family with selection headroom, and \texttt{wild}, the family with the most repair
headroom, GPT-6.1 Sol and Claude Opus 5.5 keep their $\Delta^{\mathrm{r}}_\phi$
across the four prompt levels and the three stated targets (none, CPDAG, DAG;
Table~\ref{tab:h2}, Appendix~\ref{app:full}); their cells differ by less than the
intervals. Raising the token budget per turn nearly eightfold changes nothing
either (Table~\ref{tab:h1}). Only halving the wall matters, and only for
Qwen~3.8~Flash: a third of its episodes then reach the wall and its
$\Delta^{\mathrm{r}}_\phi$ on \texttt{wild} drops from .032 to .016, while the
other two models never reach even the shorter wall.
\takeaway{The residual gains are robust to the prompt, the stated target and the token budget; only a
shorter wall lowers the gain of Qwen~3.8~Flash, whose episodes are the longest.}

\section{Conclusion}
\label{sec:conclusion}

We framed agentic causal discovery as evidence-driven adaptation at several granularities.
\heuresis{} removes the names, stresses existing algorithms with six families, and
places every agent against fixed references. With its harness, GPT-6.1 Sol
surpasses the best chosen algorithm on every family and, together with Claude
Opus 5.5, restores true edges that no algorithm proposes. 

\section*{Limitations}
\label{sec:limitations}

\textbf{Simulated data.} Real data with a trusted ground-truth causal graph are
rare, and the headroom and the per-edge decomposition of \heuresis{} need an exact
one, so all instances are simulated. Whether the gains carry over to real data
remains open and can be tested as such data become available. The six families,
the graph priors and the continuous variables at $d=10$ and $n=600$ define the setting of
every number in this paper; the generator takes $d$, $n$ and the graph density as
parameters, and new families can be added.

\textbf{Models.} Agent numbers come from three models under the neutral prompt, on
32 test instances per family, with repeats on four families
(Appendix~\ref{app:harness}); the guided condition and the prompt and budget
sweeps (on 16 test instances per family) were run once per model. Each gain is
a statement about that model with this toolbox, and every row records its model,
provider and last-turn setting. Hosted LLM APIs are not deterministic, so token
counts and cost are recorded per episode.

\textbf{Reference points.} The collection of existing algorithms is of our own
construction and is released with the benchmark, and both headrooms are
re-measured whenever algorithms are added. The repair headroom is defined on PC's
deletions, so the decomposition explains a graph only through where it differs
from the output of PC. Scores are
adjacency F1 with nSHD as the secondary metric. The code-analysis tool runs in a child process with a cleaned
environment and is not a security sandbox.

\section*{Ethics Statement and Responsible NLP Checklist Items}

The data are synthetic numeric tables with no human subjects, no personal data
and no annotators; Appendix~\ref{app:datasheet} gives a datasheet. The benchmark
is intended for evaluating the statistical judgment of LLM agents on causal
discovery; it should not be used to certify a system for causal conclusions
about real populations. The code and the development instances will be publicly
released, the code under the MIT License and the data under CC BY 4.0.
Compute and API cost, the number of runs per cell and the model versions are
reported in Appendix~\ref{app:harness}. We used Claude Code for coding and for drafting the paper; the authors
verified every number against the result files.

\bibliography{references}

\appendix

\section{Proposition~\ref{prop:limit} and its Proof}
\label{app:proof}

\begin{proposition}
\label{prop:limit}
Let $\alpha$ be the level of the test that deleted a pair and $\beta$ its power against the dependence that exists.
Condition on deletion and let $p_0$ be the recorded $p$-value. For a wrongly deleted true edge the law of $p_0$ on
$(\alpha,1]$ has density $g_\beta$; for a rightly deleted non-edge it is uniform. If $g_\beta\to1$ uniformly as the
dependence vanishes, so that $\beta\downarrow\alpha$, then the likelihood ratio between the two classes tends to 1
uniformly, and every ranker that is a function of $p_0$ has AUC $\to\tfrac12$ for detecting wrong deletions.
\end{proposition}

\paragraph{Setting.}
PC deletes a pair at the first set in its enumeration whose test returns a $p$-value above $\alpha$. Condition on deletion
and write $p_0$ for the recorded $p$-value, which lies in $(\alpha,1]$. Two assumptions idealize the two classes.

\emph{(A1) Right deletions.} For a deleted non-edge, the accepting set is a true separating set, so $p_0$ is a draw from a
true null conditioned on exceeding $\alpha$: its density on $(\alpha,1]$ is $f_r(p)=1/(1-\alpha)$.

\emph{(A2) Wrong deletions.} For a deleted true edge, every test is under the alternative. Index the alternative by an
effect size $\delta\ge0$. The $p$-value of the accepting test has density $g_\delta$ on $(0,1]$, with $g_0\equiv1$, and
power $\beta_\delta=\int_0^\alpha g_\delta$. Conditioned on deletion, the density of $p_0$ on $(\alpha,1]$ is
$f_w(p)=g_\delta(p)/(1-\beta_\delta)$, because $1-\beta_\delta=\int_\alpha^1 g_\delta$. We assume $g_\delta\to1$
uniformly on $(\alpha,1]$ as $\delta\to0$.

\paragraph{Claim.}
As $\delta\to0$, $\beta_\delta\to\alpha$, the likelihood ratio $\Lambda_\delta=f_w/f_r$ satisfies
$\sup_{(\alpha,1]}|\Lambda_\delta-1|\to0$, and for every measurable score $s(p_0)$ the AUC for detecting wrong deletions
satisfies $|\mathrm{AUC}(s)-\tfrac12|\le\tfrac12\sup|\Lambda_\delta-1|$.

\paragraph{Proof.}
By uniform convergence on a bounded interval, $1-\beta_\delta=\int_\alpha^1g_\delta\to1-\alpha$, so $\beta_\delta\to\alpha$.
Then $\Lambda_\delta(p)=g_\delta(p)(1-\alpha)/(1-\beta_\delta)$ is a ratio of uniformly convergent positive functions with
limit 1, hence converges to 1 uniformly. The total variation distance between the two classes is
$\mathrm{TV}=\tfrac12\int_\alpha^1|f_w-f_r|=\tfrac12\,\mathbb{E}_{f_r}|\Lambda_\delta-1|\le\tfrac12\sup|\Lambda_\delta-1|$.
Let $X\sim f_w$, $Y\sim f_r$ and $X'\sim f_r$ be independent, and let
$h(x,y)=\mathbf 1\{s(x)>s(y)\}+\tfrac12\mathbf 1\{s(x)=s(y)\}\in[0,1]$. Then $\mathrm{AUC}(s)=\mathbb{E}\,h(X,Y)$ and, by
symmetry, $\mathbb{E}\,h(X',Y)=\tfrac12$. Because $h$ takes values in $[0,1]$,
$|\mathbb{E}\,h(X,Y)-\mathbb{E}\,h(X',Y)|\le\mathrm{TV}$. $\square$

\paragraph{Scope of the proposition.}
The result concerns rankers that are functions of $p_0$ only and approaches $\mathrm{AUC}=\tfrac12$ as power
vanishes. Signals that use other information achieve AUCs from 0.64 to 0.90 in the first campaign
(Appendix~\ref{app:signals}).
A wrong deletion also requires some test in the enumeration to miss a real dependence, which is likely exactly where
$\beta_\delta$ is small; this is the sense in which the errors and the flat evidence share one cause. Assumption (A1) is an
idealization, since a non-edge can also be deleted at a non-separating set by a test that happens to miss a dependence.

\section{Instrument verification}
\label{app:e1}

Table~\ref{tab:e1} lists the checks run on the instruments as they now stand (31 cluster nodes, about $7\times10^5$
CPU-seconds).

\begin{table*}[t]
\centering
\footnotesize
\setlength{\tabcolsep}{5pt}
\renewcommand{\arraystretch}{1.15}
\newcolumntype{Y}[1]{>{\raggedright\arraybackslash}p{#1}}
\begin{tabular}{Y{1.0in}Y{1.85in}Y{1.5in}Y{1.35in}}
\toprule
\textbf{Check} & \textbf{Reference} & \textbf{Scale} & \textbf{Result} \\
\midrule
\multicolumn{4}{@{}l}{\textit{Graph primitives}} \\
d-separation
  & two independent implementations (moralized ancestral graph; path blocking)
  & $2.16\times10^6$ random DAGs, $4.3\times10^8$ queries
  & 0 disagreements \\
DAG to CPDAG
  & brute-force enumeration of the Markov equivalence class; Chickering's edge labeling; causal-learn
  & $3.4\times10^6$ DAGs (enumeration), $5.8\times10^6$ DAGs (labeling, causal-learn)
  & 0 disagreements \\
Exhaustive $d\le6$
  & causal-learn: d-separation, CPDAG and oracle PC
  & all $3{,}811{,}355$ labeled DAGs (equals OEIS A003024)
  & 0 disagreements \\
\midrule
\multicolumn{4}{@{}l}{\textit{PC}} \\
Oracle PC
  & true CPDAG returned, first-set and union rules
  & $6\times10^5$ random DAGs plus the exhaustive set
  & 0 failures \\
Skeleton and separating sets
  & clean-headroom PC-stable; causal-learn
  & $896{,}000$ \texttt{lg} instances, $d\in\{6,\dots,30\}$, $n\in\{300,600,2000\}$
  & identical on all \\
Seed PC (union rule)
  & causal-learn \texttt{pc(stable, uc\_priority=2)}
  & same instances
  & identical CPDAG on $892{,}563$ (99.62\%); the other $3{,}437$ have a cyclic causal-learn output \\
\midrule
\multicolumn{4}{@{}l}{\textit{Scoring}} \\
Scorer
  & gadjid SHD
  & $1.2\times10^5$ graph pairs
  & 0 disagreements \\
Invariances
  & ``no repair'' equals PC; oracle PC under relabeling
  & $1{,}800$ instances
  & pass \\
\midrule
\multicolumn{4}{@{}l}{\textit{Conditional-independence tests}} \\
Determinism
  & seeded runs across 14 nodes, 8 hash seeds, 1 to all BLAS threads
  & Fisher-$z$ and kernel tests
  & Fisher-$z$ bit-identical; kernel $\log_{10}p$ within $10^{-10}$, 0 verdict flips \\
\bottomrule
\end{tabular}
\caption{Instrument verification checks and results, grouped by the component under test.}
\label{tab:e1}
\end{table*}

\section{Generator specification and datasheet}
\label{app:datasheet}

\paragraph{Motivation and intended use.} \heuresis{} measures whether an LLM agent goes beyond calling PC when it
discovers causal structure from data alone, and, in a guided condition, what it adds to the output of PC. The benchmark targets statistical adaptation in this tool setting, alongside separate questions about general causal
reasoning and coding ability.

\paragraph{Composition.} Each instance is a tuple of an anonymous data matrix $X\in\mathbb{R}^{n\times d}$, for the guided
condition the PC seed graph and the deletion log, and a hidden record of the SCM $S$ (true DAG $G$, CPDAG, per-deletion
labels). Instances per
family: 100 development and 32 test instances, all with $d=10$ and $n=600$.

\paragraph{Mechanisms.} Linear: $X_j=\sum_i w_{ij}X_i+\varepsilon_j$, $w_{ij}\sim\pm U[0.5,2]$. MLP:
$X_j=f_j(\mathrm{pa}(j))+\varepsilon_j$, $f_j$ a random one-hidden-layer MLP with 10 tanh units. MLP with input-dependent noise:
$X_j=f_j(\mathrm{pa}(j))+g_j(\mathrm{pa}(j))\,\varepsilon_j$, $g_j=\mathrm{softplus}(\mathrm{MLP})+0.2$,
$\varepsilon_j\sim N(0,1)$. The scalar maps (piecewise linear with a random knot and two signed slopes, sigmoid,
random Fourier features, quadratic, absolute value) act on $z_j=w_j^\top X_{\mathrm{pa}(j)}$ with a random
unit-norm $w_j$; in \texttt{wild} they are standardized. Sigmoid with input-dependent noise is
$\tanh(az_j)+(\mathrm{softplus}(bz_j)+0.2)\,\varepsilon_j$. Linear with $t_3$ noise replaces $\varepsilon_j$ by scaled
Student-$t_3$ noise. Noise scales: $\sigma\sim U[0.5,2]$ for the linear nodes of \texttt{lg} and \texttt{mixed};
$U[0.5,1.5]$ for MLP nodes and the linear nodes of \texttt{wild}; $U[0.15,0.35]$ for the maps of \texttt{wild};
$U[0.15,0.40]$ in \texttt{gauntlet}, whose root nodes are $N(0,1)$.

\paragraph{Collection process.} Instances are generated by the code in the repository from a seed. Graphs are
Erd\H{o}s--R\'enyi or Barab\'asi--Albert over a uniformly random causal order, with expected degree 2 by default, so
about $d$ edges. Variables are named in random order. Columns of continuous families are standardized to mean 0 and
variance 1, which removes a known shortcut of simulated data, where the variance of a variable reveals its position in
the causal order \citep{reisach2021beware}; after standardization this shortcut is at chance level. Development and test
seeds are disjoint. Every reference algorithm is chosen on development instances and reported on test instances, and we
report distributions over instances rather than single draws \citep{mirzadeh2025gsmsymbolic}.

\paragraph{Design rationale.} The families beyond \texttt{lg} are motivated by properties that real biological data are
known to have. Protein-signaling data show nonlinear dependences between the measured proteins \citep{sachs2005causal};
\texttt{anm}, \texttt{wild} and \texttt{gauntlet} model such dependences with nonlinear maps. Gene-expression data
have noise whose scale depends on the signal \citep{love2014deseq2, hafemeister2019sctransform}; \texttt{hetero} models
this by letting the noise scale depend on the parents. Each family beyond \texttt{lg} therefore stresses an assumption
of the standard algorithms that real data of this kind also fail to meet.

\paragraph{Data statement.} The data are synthetic numeric tables. There is no human language, no human subject, no
personal data and no annotator population, so there is nothing to report on speakers or annotators
\citep{bender2018data,gebru2021datasheets}.

\paragraph{Maintenance.} Frozen development instances are public. Test seeds are drawn after the tag that fixes the prompt,
tool schema, budgets, model list and analysis code; they are sealed and rotated on a published schedule.

\section{Details of the probes in Figure~\ref{fig:chain}}
\label{app:probes}

Both probes use Qwen~3.8~Flash in thinking mode at temperature 0.

\paragraph{Leakage (Figure~\ref{fig:chain}a).}
We give the model only the variable names, with no data, and ask for the skeleton and the edge directions of seven
BNLearn networks \citep{scutari2022bnrepository} (Asia, Cancer, Earthquake, Sachs, Child, Insurance, Alarm) and, for
each of the 101 Tübingen pairs with a parseable description \citep{mooij2016distinguishing}, which variable causes
which. As controls we replace the names with indices in random order and, for the networks, add eight \heuresis{}
instances, whose variables never had names. With the names, the model gets 99\% of the Tübingen directions right and
most of the BNLearn skeletons; without them it falls to chance, and on the \heuresis{} instances it is near zero.

\paragraph{Saturation on the BNLearn structures.}
We simulate linear-Gaussian data on the anonymized BNLearn structures and run PC with a Fisher-$z$ test and with the
kernel conditional-independence test KCI \citep{zhang2011kernel}, GES, and the ideal oracle that restores every true
edge PC wrongly deleted. GES alone recovers from 31\% (Insurance) to all (Cancer, Earthquake) of what the ideal oracle
gains over \pckci{}. An agent given these instances does no better than the best algorithm: Qwen~3.8~Flash finishes
on par with GES, and in 28\% of its episodes it returns the output of PC unchanged. The continuous deviations from the
linear-Gaussian case in Figure~\ref{fig:chain}b are described in \S\ref{sec:gen} and \S\ref{sec:headroom}.

\section{The router baseline}
\label{app:baselines}

The \texttt{router} is a hand-crafted rule that decides, from the data of one instance and without the family label,
whose output to submit. It has three steps.

\paragraph{Screen.} The router first runs PC with the Fisher-$z$ test at $\alpha=0.05$ and records the pairs that PC
deletes. For each deleted pair it computes the gap between the distance correlation and the absolute Pearson correlation
of the two variables, on a fixed random subsample of at most 300 rows. A gap of at least 0.15 flags the pair as
dependent in a way that a linear test cannot see.

\paragraph{Route.} If the flagged pairs make up at least a fraction $\tau=0.2$ of the deleted pairs, the router switches
to PC with the kernel-based conditional independence test (KCI). Otherwise it keeps the output of PC with the
Fisher-$z$ test.

\paragraph{Re-test.} On the KCI branch the router re-tests every pair that PC with KCI deleted, and restores a pair only
if two conditions hold: KCI rejects independence at the conditioning set that deleted the pair, with a Holm correction
across all deleted pairs of the instance, and no alternative conditioning set that the PC graph implies for the pair
returns independence at level $\alpha$. A pair is a true non-edge if some set separates it, so a test at another set can
only confirm a deletion. On the full sample, the test at the original set is the same test on the same data that deleted
the pair; it returns the same $p$-value above $\alpha$ and cannot reject. The KCI branch therefore returns the graph of PC
with KCI, and the router equals PC with KCI on the instances it routes there and PC with the Fisher-$z$ test on the
others.

\paragraph{Foundation models.} \texttt{CDFM} \citep{qiao2026cdfm} runs from its released checkpoint
(\texttt{DMIRLAB/CDFM}) and binarizes its edge logits with the threshold that the checkpoint calibrates from the logits
themselves. \texttt{TabCausal} \citep{li2026tabcausal} runs from its released checkpoint with its exponential-moving-average
weights and the default edge threshold of 0.5. Both read the same $(n,d)$ data matrix as the agents in one forward pass; nothing is tuned on
\heuresis, and neither sees the family label or a development split.

\section{Repairing the output of PC from its own log}
\label{app:signals}

\begin{figure}[t]
\centering
\includegraphics[width=\columnwidth]{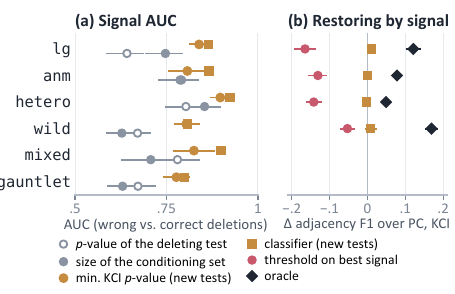}
\caption{Scripted use of the deletion log of \pckci{}. (a) AUC of each signal for ranking the wrongly deleted edges
among all deletions; the first two signals need no new tests. (b) Change in adjacency F1 from restoring edges by each
signal, compared with the ideal oracle that restores every wrongly deleted edge. 32 test instances per family. Error bars
represent bootstrapped 95\% confidence intervals.}
\label{fig:signals}
\end{figure}

A script can also try to find PC's mistakes without a model of the data. Figure~\ref{fig:signals}a ranks the signals that
the deletion log of \pckci{} offers for telling a wrongly deleted edge from a rightly deleted one. The two signals
available from the log alone, the $p$-value of the deleting test and the size of its conditioning set, rank wrong
deletions only moderately, and a classifier on the four such features restores nothing at any threshold chosen on
development instances. For the $p$-value this is what Proposition~\ref{prop:limit} predicts where the deleting test has
little power: a ranker that depends on the recorded $p$-value alone tends to chance as power vanishes. Signals that need new tests rank better: the minimum KCI $p$-value over the alternative
conditioning sets that the PC graph implies for the pair is the best single signal on every family, and a
gradient-boosted classifier on all nine features, trained with the family held out, ranks higher still. Ranking the
deletions, however, does not translate into a gain from restoring them (Figure~\ref{fig:signals}b). Because wrong
deletions are only a few percent of all deletions, a threshold on the best signal restores far more right deletions
than wrong ones and lowers adjacency F1 on all four families it was run on (\texttt{lg}, \texttt{anm}, \texttt{hetero}
and \texttt{wild}), and the classifier at its best threshold gains nothing distinguishable from zero.

\section{Reference scores, agent results and the guided condition in full}
\label{app:full}

Table~\ref{tab:headroom} gives the reference scores and the two kinds of headroom behind Figure~\ref{fig:chain}c and
\S\ref{sec:recoverable}; Table~\ref{tab:m1} the agent results behind Figure~\ref{fig:agents}, with every
repeat and last-turn setting in Table~\ref{tab:m1rep}; and Table~\ref{tab:guided} the guided condition behind
Figure~\ref{fig:guided}.

\begin{table*}[p]
\centering
\footnotesize
\setlength{\tabcolsep}{5pt}
\begin{tabular}{lrrrrrr}
\toprule
 & \multicolumn{2}{c}{GPT-6.1 Sol} & \multicolumn{2}{c}{Claude Opus 5.5} & \multicolumn{2}{c}{Qwen~3.8~Flash} \\
\cmidrule(lr){2-3}\cmidrule(lr){4-5}\cmidrule(lr){6-7}
 & F1 & $\Delta^{\mathrm{r}}_\phi$ & F1 & $\Delta^{\mathrm{r}}_\phi$ & F1 & $\Delta^{\mathrm{r}}_\phi$ \\
\midrule
\texttt{lg}       & .978 & \textbf{.043} [.019, .070] & .972 & \textbf{.037} [.013, .066] & .945 & .009 [$-$.018, .039] \\
\texttt{anm}      & .958 & \textbf{.052} [.032, .074] & .906 & $-$.000 [$-$.027, .027] & .899 & $-$.008 [$-$.042, .027] \\
\texttt{hetero}   & .967 & \textbf{.042} [.027, .058] & .934 & .010 [$-$.012, .031] & .903 & $-$.022 [$-$.048, .001] \\
\texttt{wild}     & .935 & \textbf{.115} [.083, .146] & .880 & \textbf{.060} [.030, .090] & .843 & .023 [$-$.013, .059] \\
\texttt{mixed}    & .972 & \textbf{.045} [.028, .065] & .952 & .025 [$-$.000, .051] & .924 & $-$.003 [$-$.033, .028] \\
\texttt{gauntlet} & .933 & \textbf{.084} [.059, .111] & .894 & \textbf{.046} [.027, .063] & .871 & .023 [$-$.000, .046] \\
\midrule
$\Delta^{\mathrm{s}}_\phi$ (all)          & \multicolumn{2}{c}{\textbf{.075} [.064, .087]} & \multicolumn{2}{c}{\textbf{.041} [.029, .054]} & \multicolumn{2}{c}{\textbf{.015} [.003, .029]} \\
$\Delta^{\mathrm{r}}_\phi$ (all) & \multicolumn{2}{c}{\textbf{.064} [.053, .074]} & \multicolumn{2}{c}{\textbf{.030} [.020, .040]} & \multicolumn{2}{c}{.004 [$-$.009, .016]} \\
\bottomrule
\end{tabular}
\caption{Agent performance (every episode submitted a graph) on 32 test instances per family: adjacency F1
and the paired difference from the family's best chosen algorithm. Rows marked (all) are over all families. GPT-6.1 Sol and Claude Opus 5.5 are shown in repeat 1; their repeats 2 and 3, and the
finished-only and all-episode views of the runs without a retry in the last turn, are in Table~\ref{tab:m1rep}.}
\label{tab:m1}
\end{table*}

\begin{table*}[p]
\centering
\footnotesize
\setlength{\tabcolsep}{5pt}
\begin{tabular}{@{}l rcc @{\hspace{14pt}} rcc@{}}
\toprule
 & \multicolumn{3}{c}{GPT-6.1 Sol} & \multicolumn{3}{c}{Claude Opus 5.5} \\
\cmidrule(lr){2-4}\cmidrule(l){5-7}
Family & $n$ & adjacency F1 & nSHD & $n$ & adjacency F1 & nSHD \\
\midrule
\texttt{lg} & 32 & \textbf{.125} [.086, .166] & \textbf{.068} [.047, .091] & 23 & \textbf{.134} [.080, .189] & \textbf{.079} [.050, .108] \\
\texttt{anm} & 32 & \textbf{.071} [.047, .097] & \textbf{.110} [.091, .128] & 30 & .018 [$-$.014, .048] & \textbf{.088} [.065, .111] \\
\texttt{hetero} & 32 & \textbf{.053} [.035, .072] & \textbf{.086} [.069, .103] & 26 & \textbf{.027} [.002, .053] & \textbf{.039} [.014, .062] \\
\texttt{wild} & 32 & \textbf{.122} [.091, .152] & \textbf{.135} [.115, .156] & 30 & \textbf{.068} [.039, .097] & \textbf{.111} [.090, .131] \\
\texttt{mixed} & 32 & \textbf{.053} [.035, .071] & \textbf{.092} [.069, .114] & 30 & \textbf{.035} [.007, .064] & \textbf{.077} [.053, .101] \\
\texttt{gauntlet} & 32 & \textbf{.097} [.066, .133] & \textbf{.133} [.117, .149] & 26 & \textbf{.062} [.033, .094] & \textbf{.109} [.091, .126] \\
\midrule
\texttt{lg}, vs GES & 32 & \textbf{.043} [.019, .072] & \textbf{.027} [.008, .050] & 20 & \textbf{.055} [.017, .098] & \textbf{.043} [.016, .076] \\
\bottomrule
\end{tabular}
\caption{Agent gain over the in-episode output of PC-KCI (GES on the last row of each model), scored by the same scorer on the same instance, repeat 1, mean paired gain with 95\% bootstrap intervals over instances; positive is better for the agent, and nSHD is reported as its reduction. $n$ is the number of episodes in which the agent called the algorithm (all 32 for GPT-6.1 Sol). Bold: the interval excludes zero.}
\label{tab:nshd}
\end{table*}

\begin{table*}[p]
\centering
\footnotesize
\setlength{\tabcolsep}{5pt}
\renewcommand{\arraystretch}{0.92}
\begin{tabular}{llrrrr}
\toprule
Toolbox & Family & Precision & Recall & vs \pckci{} & Guided $-$ unguided \\
\midrule
\multicolumn{6}{l}{GPT-6.1 Sol} \\
full & \texttt{lg} & .88 [.77, .96] & .94 [.89, .99] & .112 [.074, .149] & $-$.006 [$-$.019, .004] \\
 & \texttt{anm} & .87 [.78, .95] & .76 [.63, .88] & .048 [.028, .072] & $-$.003 [$-$.016, .008] \\
 & \texttt{hetero} & .89 [.79, .98] & .80 [.68, .91] & .039 [.023, .055] & $-$.004 [$-$.013, .006] \\
 & \texttt{wild} & .98 [.94, 1.00] & .70 [.59, .80] & .125 [.093, .156] & .009 [$-$.007, .028] \\
\midrule
minimal & \texttt{lg} & .83 [.73, .93] & .90 [.84, .96] & .091 [.057, .125] & $-$.027 [$-$.051, $-$.006] \\
 & \texttt{anm} & .62 [.47, .78] & .57 [.42, .72] & .005 [$-$.023, .033] & $-$.047 [$-$.075, $-$.021] \\
 & \texttt{hetero} & .81 [.68, .92] & .61 [.45, .75] & .014 [$-$.001, .028] & $-$.029 [$-$.043, $-$.015] \\
 & \texttt{wild} & .83 [.71, .93] & .47 [.34, .61] & .063 [.032, .093] & $-$.053 [$-$.075, $-$.033] \\
\midrule
\multicolumn{6}{l}{Claude Opus 5.5} \\
full & \texttt{lg} & .89 [.80, .96] & .94 [.89, .99] & .109 [.077, .144] & $-$.003 [$-$.020, .013] \\
 & \texttt{anm} & .66 [.51, .80] & .55 [.39, .69] & .009 [$-$.018, .034] & .009 [$-$.011, .029] \\
 & \texttt{hetero} & .75 [.61, .88] & .64 [.47, .79] & .007 [$-$.015, .028] & $-$.003 [$-$.016, .012] \\
 & \texttt{wild} & .86 [.76, .95] & .49 [.37, .62] & .070 [.043, .097] & .009 [$-$.013, .034] \\
\midrule
minimal & \texttt{lg} & .86 [.77, .95] & .88 [.79, .95] & .095 [.061, .131] & $-$.017 [$-$.030, $-$.004] \\
 & \texttt{anm} & .42 [.26, .58] & .54 [.39, .69] & $-$.031 [$-$.065, .007] & $-$.030 [$-$.059, $-$.001] \\
 & \texttt{hetero} & .62 [.45, .78] & .60 [.43, .75] & $-$.010 [$-$.032, .012] & $-$.020 [$-$.038, $-$.002] \\
 & \texttt{wild} & .50 [.35, .63] & .41 [.27, .54] & .021 [$-$.006, .050] & $-$.040 [$-$.076, $-$.009] \\
\midrule
\multicolumn{6}{l}{Qwen~3.8~Flash} \\
full & \texttt{lg} & .89 [.80, .97] & .91 [.81, .98] & .096 [.046, .141] & .012 [$-$.035, .053] \\
 & \texttt{anm} & .71 [.56, .85] & .42 [.27, .58] & .004 [$-$.018, .026] & .012 [$-$.020, .045] \\
 & \texttt{hetero} & .87 [.75, .97] & .44 [.27, .60] & $-$.003 [$-$.018, .013] & .020 [$-$.011, .052] \\
 & \texttt{wild} & .78 [.65, .90] & .35 [.22, .49] & .033 [.012, .056] & .009 [$-$.024, .043] \\
\midrule
minimal & \texttt{lg} & .84 [.71, .94] & .56 [.42, .71] & .034 [$-$.002, .071] & $-$.050 [$-$.084, $-$.018] \\
 & \texttt{anm} & .56 [.40, .72] & .35 [.20, .51] & $-$.034 [$-$.056, $-$.010] & $-$.026 [$-$.063, .011] \\
 & \texttt{hetero} & .85 [.74, .95] & .56 [.39, .72] & .008 [$-$.004, .020] & .030 [.004, .060] \\
 & \texttt{wild} & .71 [.55, .85] & .23 [.12, .35] & $-$.002 [$-$.029, .023] & $-$.025 [$-$.059, .008] \\
\bottomrule
\end{tabular}
\caption{Guided condition (32 test instances per family, all episodes): precision and recall of restoring the wrongly deleted edges of \pckci{}, the paired gain over \pckci{}, and the paired
gain over the agent's own unguided run with the full toolbox on the same instances (repeat 1 for GPT-6.1 Sol and
Claude Opus 5.5), with 95\% bootstrap intervals over episodes.}
\label{tab:guided}
\end{table*}

\begin{table*}[p]
\centering
\footnotesize
\setlength{\tabcolsep}{5pt}
\renewcommand{\arraystretch}{0.92}
\begin{tabular}{llrrrrr}
\toprule
 & & \multicolumn{2}{c}{\texttt{lg}} & \multicolumn{2}{c}{\texttt{wild}} & \\
\cmidrule(lr){3-4}\cmidrule(lr){5-6}
Factor & Level & $\Delta^{\mathrm{r}}_\phi$ & shortcut & $\Delta^{\mathrm{r}}_\phi$ & shortcut & wall \\
\midrule
\multicolumn{7}{l}{GPT-6.1 Sol} \\
prompt & neutral & .049 [.028, .075] & .21 & .101 [.078, .125] & .00 & .00 \\
 & scoring & .050 [.029, .076] & .23 & .104 [.081, .128] & .02 & .00 \\
 & warning & .053 [.030, .079] & .19 & .095 [.076, .115] & .00 & .00 \\
 & rule & .044 [.021, .070] & .12 & .104 [.083, .125] & .02 & .00 \\
\midrule
target & none & .049 [.030, .069] & .20 & .103 [.084, .123] & .00 & .00 \\
 & CPDAG & .049 [.031, .072] & .20 & .095 [.077, .113] & .03 & .00 \\
 & DAG & .050 [.030, .072] & .16 & .106 [.087, .125] & .00 & .00 \\
\midrule
\multicolumn{7}{l}{Claude Opus 5.5} \\
prompt & neutral & .023 [.000, .047] & .19 & .056 [.036, .078] & .00 & .00 \\
 & scoring & .039 [.016, .064] & .15 & .065 [.041, .087] & .02 & .00 \\
 & warning & .036 [.012, .063] & .12 & .046 [.024, .068] & .02 & .00 \\
 & rule & .021 [$-$.005, .048] & .15 & .056 [.035, .077] & .04 & .00 \\
\midrule
target & none & .032 [.013, .054] & .16 & .056 [.038, .074] & .00 & .00 \\
 & CPDAG & .023 [.001, .046] & .20 & .056 [.036, .075] & .06 & .00 \\
 & DAG & .034 [.013, .058] & .09 & .055 [.037, .074] & .00 & .00 \\
\midrule
\multicolumn{7}{l}{Qwen~3.8~Flash} \\
prompt & neutral & .005 [$-$.038, .042] & .12 & .031 [.006, .055] & .00 & .08 \\
 & scoring & .031 [.005, .060] & .08 & .019 [$-$.007, .044] & .00 & .10 \\
 & warning & .019 [$-$.010, .048] & .10 & .019 [$-$.002, .040] & .00 & .11 \\
 & rule & $-$.027 [$-$.081, .021] & .06 & .026 [.008, .045] & .00 & .23 \\
\midrule
target & none & .007 [$-$.030, .041] & .09 & .020 [.001, .037] & .00 & .14 \\
 & CPDAG & .004 [$-$.029, .035] & .19 & .018 [$-$.004, .039] & .00 & .14 \\
 & DAG & .010 [$-$.025, .041] & .00 & .033 [.014, .052] & .00 & .12 \\
\bottomrule
\end{tabular}
\caption{Prompt level (each level adds to the previous one: neutral task statement; budgets and the scoring statement; a warning that standard algorithms can fail under nonlinear or heteroscedastic mechanisms; PC's deletion rule) and stated target (first 16 of the 32 test instances per family, each
row averaged over the other factor; GPT-6.1 Sol and Claude Opus 5.5 in repeat 1): paired gain in adjacency F1 over the best chosen algorithm, share of episodes whose
final graph is PC's output, and share of episodes that reached the wall.}
\label{tab:h2}
\end{table*}

\begin{table*}[p]
\centering
\footnotesize
\setlength{\tabcolsep}{5pt}
\renewcommand{\arraystretch}{0.92}
\begin{tabular}{llrrrrr}
\toprule
Factor & Level & $\Delta^{\mathrm{r}}_\phi$, \texttt{lg} & $\Delta^{\mathrm{r}}_\phi$, \texttt{wild} & wall & unsubmitted & tokens \\
\midrule
\multicolumn{7}{l}{GPT-6.1 Sol} \\
tokens per turn & 4\,096 & .044 [.019, .074] & .112 [.084, .141] & .00 & .00 & 7 k \\
 & 8\,000 & .043 [.017, .072] & .094 [.070, .121] & .00 & .00 & 7 k \\
 & 16\,000 & .051 [.026, .080] & .099 [.075, .123] & .00 & .00 & 6 k \\
 & 32\,000 & .045 [.016, .076] & .109 [.080, .142] & .00 & .00 & 6 k \\
\midrule
wall & 1\,200 s & .049 [.030, .071] & .107 [.088, .128] & .00 & .00 & 7 k \\
 & 2\,400 s & .042 [.023, .064] & .100 [.082, .119] & .00 & .00 & 7 k \\
\midrule
\multicolumn{7}{l}{Claude Opus 5.5} \\
tokens per turn & 4\,096 & .029 [.002, .061] & .054 [.026, .079] & .00 & .00 & 11 k \\
 & 8\,000 & .027 [$-$.001, .059] & .038 [.013, .063] & .00 & .00 & 11 k \\
 & 16\,000 & .034 [.007, .064] & .051 [.026, .076] & .00 & .00 & 11 k \\
 & 32\,000 & .028 [$-$.001, .059] & .048 [.025, .070] & .00 & .00 & 11 k \\
\midrule
wall & 1\,200 s & .033 [.013, .055] & .054 [.036, .073] & .00 & .00 & 11 k \\
 & 2\,400 s & .026 [.006, .048] & .041 [.023, .058] & .00 & .00 & 11 k \\
\midrule
\multicolumn{7}{l}{Qwen~3.8~Flash} \\
tokens per turn & 4\,096 & .011 [$-$.014, .040] & .020 [$-$.011, .048] & .19 & .02 & 106 k \\
 & 8\,000 & $-$.017 [$-$.069, .031] & .024 [$-$.018, .062] & .25 & .02 & 101 k \\
 & 16\,000 & .007 [$-$.027, .045] & .025 [$-$.001, .051] & .22 & .02 & 99 k \\
 & 32\,000 & .003 [$-$.051, .050] & .027 [.007, .045] & .22 & .02 & 101 k \\
\midrule
wall & 1\,200 s & $-$.003 [$-$.035, .028] & .016 [$-$.008, .038] & .34 & .02 & 70 k \\
 & 2\,400 s & .005 [$-$.024, .032] & .032 [.013, .051] & .09 & .02 & 134 k \\
\bottomrule
\end{tabular}
\caption{Budget sweep (first 16 of the 32 test instances per family, each row averaged over
the other factor; GPT-6.1 Sol and Claude Opus 5.5 in repeat 1): paired gain in adjacency F1 over the best chosen algorithm, share of episodes that reached the wall and
submitted in the retry, share left unsubmitted, and completion tokens per episode.}
\label{tab:h1}
\end{table*}

\begin{table}[t]
\centering
\footnotesize
\setlength{\tabcolsep}{3pt}
\resizebox{\columnwidth}{!}{%
\begin{tabular}{lrrrrr}
\toprule
Family & empty & complete & \pckci{} & random, oracle's count & oracle \\
\midrule
\texttt{lg}       & .000 & .361 & .861 & .811 [.773, .848] & .970 \\
\texttt{anm}      & .000 & .363 & .906 & .857 [.830, .884] & .984 \\
\texttt{hetero}   & .000 & .362 & .928 & .892 [.871, .913] & .981 \\
\texttt{wild}     & .000 & .362 & .818 & .742 [.710, .774] & .973 \\
\texttt{mixed}    & .000 & .378 & .934 & .899 [.867, .928] & .987 \\
\texttt{gauntlet} & .000 & .340 & .849 & .768 [.738, .799] & .990 \\
\bottomrule
\end{tabular}}
\caption{Trivial floors (adjacency F1, 32 test instances per family): the empty graph, the complete graph (every deletion of \pckci{} restored), \pckci{} itself
(nothing restored), a random restoration of as many deleted pairs as the oracle restores (20 draws per instance), and
the oracle.}
\label{tab:floors}
\end{table}

\section{Harness and reporting disclosure}
\label{app:harness}

Table~\ref{tab:harness} records, for every agent result in this paper, the model, harness and budget it was obtained under.

\begin{figure}[t]
\centering
\includegraphics[width=\columnwidth]{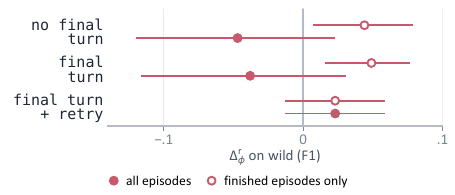}
\caption{Qwen~3.8~Flash on \texttt{wild} under the three last-turn settings of Table~\ref{tab:harness} (no last turn,
one last turn in which it can only submit, the same turn with a retry), scored over all episodes and over
finished episodes only. Intervals as in Figure~\ref{fig:agents}.}
\label{fig:truncation}
\end{figure}

\paragraph{Truncated episodes.}
\label{app:truncation}
Without a retry in the last turn, some episodes reached the time limit without submitting a graph, up to a fifth of them on
\texttt{wild} (Table~\ref{tab:m1rep}). Over the finished episodes alone, Qwen~3.8~Flash scored clearly above \pckci{}
on \texttt{wild}, a family without selection headroom, where any gain would lie above the best chosen algorithm; over all episodes
it scored below (Figure~\ref{fig:truncation}). The episodes that ran out of budget were the hard ones; conditioning on completion therefore inflates the score.
The last turn therefore reads the final edge list tolerantly and allows one retry after a rejected list, so that every episode submits a
graph and the two views coincide. We therefore fix the following reporting rule for every later
run:
every cell is reported both over all episodes and over finished episodes, with the truncation rate beside them, and
we claim a capability only under a harness in which the two agree.

\begin{table*}[p]
\centering
\footnotesize
\setlength{\tabcolsep}{3pt}
\begin{tabular}{p{1.0in}p{5.0in}}
\toprule
Item & Value \\
\midrule
Models & GPT-6.1 Sol (OpenAI), reasoning effort ``medium''; Claude Opus 5.5 (Anthropic), adaptive thinking at effort ``high''; Qwen~3.8~Flash (Alibaba), thinking mode on. Main runs of 2026-10-09; guided, prompt and budget runs of GPT-6.1 Sol and Claude Opus 5.5 of 2026-10-10 \\
Prompt & neutral level, no stated target; no algorithm is named below the rule level (Table~\ref{tab:h2}) \\
Toolbox & the full toolbox: algorithm menu, conditional-independence tests (single, batched and subsampling stability), code analysis, local score, orientation probes, data summary, graph editor, and submission of the final edge list \\
Budgets & 1000 CI tests; 60 turns; 16000 output tokens per turn; 2400 s wall; 12 menu calls; all stated to the model; the tokens per turn and the wall are swept in Table~\ref{tab:h1} \\
Last-turn setting & no last turn (repeat 1); one last turn in which the agent can only submit (repeat 2); the same turn with tolerant reading of the edge list and one retry (later runs) \\
Runs per cell & 32 test instances per family. GPT-6.1 Sol and Claude Opus 5.5: repeats 1 to 3 on \texttt{lg}, \texttt{anm}, \texttt{hetero} and \texttt{wild}, repeat 1 on \texttt{mixed} and \texttt{gauntlet}; Qwen~3.8~Flash: the same cells as the other two models in Table~\ref{tab:m1rep}. Guided, prompt and budget runs: repeat 1 for every model (32 instances per family guided, 16 per family in the prompt and budget sweeps) \\
Episode profile & GPT-6.1 Sol (448 episodes): median 17 turns (quartiles 11 and 23); 0.39 M prompt tokens; 9.4 k completion tokens of which 4.5 k reasoning; median 4.9 min wall (maximum 21.0). Claude Opus 5.5 (all 448 episodes): median 15 turns (quartiles 10 and 22); 0.39 M prompt tokens; 14.9 k completion tokens of which 8.2 k reasoning; median 3.3 min wall (maximum 14.7). Qwen~3.8~Flash: 22 to 35 turns; 0.5 to 0.8 M prompt tokens; 140 k completion tokens of which about 125 k reasoning; 30 to 40 min wall \\
Cost per episode & GPT-6.1 Sol: about \$0.19 mean, about \$86 for its 448 episodes. Claude Opus 5.5: \$1.60 mean (median 1.06, maximum 10.64), \$718 for its 448 episodes (\$204 for the 128 of repeat 1 on four families). Qwen~3.8~Flash: \$0.085 mean (median 0.089, maximum 0.129); the main sweep of 256 episodes cost \$22 \\
Compute and software & The agents call hosted model APIs; their parameter counts are not public. Reference algorithms and instrument checks ran on CPU cluster nodes (AMD EPYC 9754); inference of \texttt{CDFM} and \texttt{TabCausal} took a few minutes on one NVIDIA H800 GPU. Python with causal-learn 0.1.4.8 (\texttt{pc} with \texttt{stable=True}, \texttt{uc\_rule=0}, \texttt{uc\_priority=2}); \texttt{CDFM} checkpoint revision \texttt{fd3ac4f} \\
\bottomrule
\end{tabular}
\caption{Harness disclosure for the agent results.}
\label{tab:harness}
\end{table*}

Table~\ref{tab:m1rep} gives every repeat of the main condition separately, with the last-turn setting it ran under, the
number of episodes that reached the wall and the number of those that still submitted a graph in the last
turn, so that the all-episode and finished-only views of Table~\ref{tab:m1} can be traced to
the change in that turn that separates them.

\begin{table*}[p]
\centering
\footnotesize
\setlength{\tabcolsep}{4pt}
\begin{tabular}{llcccrrrrr}
\toprule
 & & & & \multicolumn{2}{c}{all episodes} & \multicolumn{3}{c}{finished} \\
\cmidrule(lr){5-6}\cmidrule(lr){7-9}
Family & Repeat & Last turn & Wall / rescued & F1 & $\Delta^{\mathrm{r}}_\phi$ & $n$ & F1 & $\Delta^{\mathrm{r}}_\phi$ \\
\midrule
\multicolumn{9}{l}{GPT-6.1 Sol (reasoning effort ``medium'')} \\
\texttt{lg}     & 1 & retry & 0 / 0 & .978 & .043 [.019, .070] & 32 & .978 & .043 [.019, .070] \\
                & 2 & retry & 0 / 0 & .983 & .048 [.021, .078] & 32 & .983 & .048 [.021, .078] \\
                & 3 & retry & 0 / 0 & .983 & .048 [.021, .077] & 32 & .983 & .048 [.021, .077] \\
\texttt{anm}    & 1 & retry & 0 / 0 & .958 & .052 [.032, .074] & 32 & .958 & .052 [.032, .074] \\
                & 2 & retry & 0 / 0 & .956 & .050 [.029, .072] & 32 & .956 & .050 [.029, .072] \\
                & 3 & retry & 0 / 0 & .946 & .040 [.008, .068] & 32 & .946 & .040 [.008, .068] \\
\texttt{hetero} & 1 & retry & 0 / 0 & .967 & .042 [.027, .058] & 32 & .967 & .042 [.027, .058] \\
                & 2 & retry & 0 / 0 & .960 & .036 [.018, .053] & 32 & .960 & .036 [.018, .053] \\
                & 3 & retry & 0 / 0 & .961 & .037 [.019, .055] & 32 & .961 & .037 [.019, .055] \\
\texttt{wild}   & 1 & retry & 0 / 0 & .935 & .115 [.083, .146] & 32 & .935 & .115 [.083, .146] \\
                & 2 & retry & 0 / 0 & .939 & .119 [.089, .150] & 32 & .939 & .119 [.089, .150] \\
                & 3 & retry & 0 / 0 & .933 & .113 [.083, .142] & 32 & .933 & .113 [.083, .142] \\
\texttt{mixed}  & 1 & retry & 0 / 0 & .972 & .045 [.028, .065] & 32 & .972 & .045 [.028, .065] \\
\texttt{gauntlet} & 1 & retry & 0 / 0 & .933 & .084 [.059, .111] & 32 & .933 & .084 [.059, .111] \\
\midrule
\multicolumn{9}{l}{Claude Opus 5.5 (adaptive thinking, effort ``high'')} \\
\texttt{lg}     & 1 & retry & 0 / 0 & .972 & .037 [.013, .066] & 32 & .972 & .037 [.013, .066] \\
                & 2 & retry & 0 / 0 & .966 & .031 [.006, .060] & 32 & .966 & .031 [.006, .060] \\
                & 3 & retry & 0 / 0 & .975 & .040 [.014, .068] & 32 & .975 & .040 [.014, .068] \\
\texttt{anm}    & 1 & retry & 0 / 0 & .906 & $-$.000 [$-$.027, .027] & 32 & .906 & $-$.000 [$-$.027, .027] \\
                & 2 & retry & 0 / 0 & .914 & .008 [$-$.023, .035] & 32 & .914 & .008 [$-$.023, .035] \\
                & 3 & retry & 0 / 0 & .912 & .006 [$-$.025, .037] & 32 & .912 & .006 [$-$.025, .037] \\
\texttt{hetero} & 1 & retry & 0 / 0 & .934 & .010 [$-$.012, .031] & 32 & .934 & .010 [$-$.012, .031] \\
                & 2 & retry & 0 / 0 & .934 & .009 [$-$.011, .028] & 32 & .934 & .009 [$-$.011, .028] \\
                & 3 & retry & 0 / 0 & .930 & .005 [$-$.021, .028] & 32 & .930 & .005 [$-$.021, .028] \\
\texttt{wild}   & 1 & retry & 0 / 0 & .880 & .060 [.030, .090] & 32 & .880 & .060 [.030, .090] \\
                & 2 & retry & 0 / 0 & .898 & .078 [.046, .111] & 32 & .898 & .078 [.046, .111] \\
                & 3 & retry & 0 / 0 & .899 & .079 [.048, .111] & 32 & .899 & .079 [.048, .111] \\
\texttt{mixed}  & 1 & retry & 0 / 0 & .952 & .025 [$-$.000, .051] & 32 & .952 & .025 [$-$.000, .051] \\
\texttt{gauntlet} & 1 & retry & 0 / 0 & .894 & .046 [.027, .063] & 32 & .894 & .046 [.027, .063] \\
\midrule
\multicolumn{9}{l}{Qwen~3.8~Flash (thinking mode)} \\
\texttt{lg}     & 1 & none & 2 / 0 & .935 & $-$.000 [$-$.046, .039] & 30 & .966 & .024 [$-$.003, .052] \\
                & 2 & once & 2 / 1 & .944 & .008 [$-$.043, .050] & 31 & .962 & .028 [$-$.002, .059] \\
                & 3 & retry & 2 / 2 & .945 & .009 [$-$.018, .039] & 32 & .945 & .009 [$-$.018, .039] \\
\texttt{anm}    & 1 & retry & 5 / 5 & .899 & $-$.008 [$-$.042, .027] & 32 & .899 & $-$.008 [$-$.042, .027] \\
                & 2 & retry & 2 / 2 & .890 & $-$.016 [$-$.047, .015] & 32 & .890 & $-$.016 [$-$.047, .015] \\
                & 3 & retry & 4 / 4 & .881 & $-$.026 [$-$.064, .013] & 32 & .881 & $-$.026 [$-$.064, .013] \\
\texttt{hetero} & 1 & retry & 4 / 4 & .903 & $-$.022 [$-$.048, .001] & 32 & .903 & $-$.022 [$-$.048, .001] \\
                & 2 & retry & 5 / 5 & .889 & $-$.035 [$-$.071, $-$.002] & 32 & .889 & $-$.035 [$-$.071, $-$.002] \\
                & 3 & retry & 0 / 0 & .880 & $-$.044 [$-$.080, $-$.014] & 32 & .880 & $-$.044 [$-$.080, $-$.014] \\
\texttt{wild}   & 1 & none & 6 / 0 & .773 & $-$.047 [$-$.120, .023] & 26 & .865 & .044 [.007, .079] \\
                & 2 & once & 5 / 0 & .782 & $-$.038 [$-$.116, .031] & 27 & .865 & .049 [.016, .077] \\
                & 3 & retry & 2 / 2 & .843 & .023 [$-$.013, .059] & 32 & .843 & .023 [$-$.013, .059] \\
\texttt{mixed}  & 1 & retry & 9 / 9 & .924 & $-$.003 [$-$.033, .028] & 32 & .924 & $-$.003 [$-$.033, .028] \\
\texttt{gauntlet} & 1 & retry & 5 / 5 & .871 & .023 [$-$.000, .046] & 32 & .871 & .023 [$-$.000, .046] \\
\bottomrule
\end{tabular}
\caption{Per-repeat results for the main agent condition. F1 is adjacency F1; $\Delta^{\mathrm{r}}_\phi$ is the paired difference from
the family's best chosen algorithm. Repeats 1 to 3 on four families and repeat 1 on \texttt{mixed} and \texttt{gauntlet}, the same for the three models. Last turn: none, one turn in which the agent can only submit, or that turn with a retry.}
\label{tab:m1rep}
\end{table*}

\end{document}